\pdfoutput=1

\documentclass[11pt]{article}
\usepackage{url}
\usepackage[]{ACL2023}
\usepackage{cuted}
\usepackage{times}
\usepackage{latexsym}
\usepackage{tcolorbox}
\usepackage[T1]{fontenc}

\usepackage[utf8]{inputenc}

\usepackage{microtype}

\usepackage{inconsolata}
\usepackage{enumitem}
\usepackage{xcolor}
\title{
Assessing the Effectiveness of GPT-4o in Climate Change Evidence Synthesis and Systematic Assessments: Preliminary Insights}

\author{Elphin Tom Joe \and
  Sai Dileep Koneru \and
  Christine J Kirchhoff\\
  Pennsylvania State University\\
  \{etj5074, sdk96, cxk475\}@psu.edu
  }

\begin{document}
\maketitle
\begin{abstract}
In this research short, we examine the potential of using GPT-4o, a state-of-the-art large language model (LLM) to undertake evidence synthesis and systematic assessment tasks. Traditional workflows for such tasks involve large groups of domain experts who manually review and synthesize vast amounts of literature. The exponential growth of scientific literature and recent advances in LLMs provide an opportunity to complementing these traditional workflows with new age tools. We assess the efficacy of GPT-4o to do these tasks on a sample from the dataset created by the Global Adaptation Mapping Initiative (GAMI) where we check the accuracy of climate change adaptation related feature extraction from the scientific literature across three levels of expertise. Our results indicate that while GPT-4o can achieve high accuracy in low-expertise tasks like geographic location identification, their performance in intermediate and high-expertise tasks, such as stakeholder identification and assessment of depth of the adaptation response, is less reliable. The findings motivate the need for designing assessment workflows that utilize the strengths of models like GPT-4o while also providing refinements to improve their performance on these tasks.
\end{abstract}

\section{Introduction}

Climate change is one of the most pressing challenges that several regions across the world have to face in the coming decades \cite{lee2023ipcc}. Adapting to climate change is essential for ensuring long-term sustainability \cite{styczynski2014decision}. For decision-makers to effectively respond to this challenge, they must carefully plan their strategies based on well-documented and assessed climate adaptation evidence. This involves reviewing a vast array of scientific documents and case studies that detail adaptation efforts in different regions. The Intergovernmental Panel on Climate Change (IPCC) has formalized the assessment of this evidence through the publication of time-sensitive reports. These reports play a crucial role in informing international treaties and country-specific legislative actions.


While traditionally such assessments relied on domain experts working voluntarily in teams tasked with annotating the documents for specific aspects of climate change, this is changing of late with the incorporation of machine learning in the evidence gathering and synthesis process \cite{berrang2021systematic, sietsma2024next}. This is because the exponential growth of scientific literature over time has made the process of managing and synthesizing the evidence increasingly challenging \cite{bornmann2021growth}. Automating the annotation of large volumes of scientific data can save valuable researcher time and reduce the assessment cycle, allowing decision-makers to receive quicker and more up-to-date information. 

Recent advancements in neural network methods have shown to be very useful in processing documents and extracting useful information from documents in open domain data such as Wikipedia articles \cite{martinez2020information}. However, scientific documents present unique challenges due to their complex domain-specific terminologies and concepts. Addressing these challenges requires training models with high-quality, human-labeled data at a sufficient scale. Recent advances in Large Language Models (LLMs) have shown promise in overcoming these challenges. Their diverse training across various topics has made them effective at extracting information from scientific documents and supporting researchers in enhancing time efficiency. Although they perform well in extracting information from scientific documents in certain domains, they still struggle to fully understand the evidence presented in these articles \cite{koneru2023can}. 

Given the importance of accurate information extraction for decision-making, it is crucial to have a reliable model that ensures factual accuracy during the extraction phase. In this preliminary work, we explore the potential of using GPT-4o, a state-of-the-art LLM by OpenAI, without explicit domain specific training, as extractors of useful information from documents related to climate adaptation. Our contributions are as follows 
\begin{itemize}[leftmargin=*]
    \item We empirically evaluate the utility of GPT-4o for annotating climate-related texts for systematic assessments. This evaluation involves comparing the performance of GPT-4o to human annotators. 
    \item We evaluate the annotation capability across different levels of information complexity. For this, we test the extraction of features at varying levels of expertise: low, medium, and high. For low-level information, we identify direct feature that can be extracted without any domain expertise. Medium-level feature requires the model to use a taxonomy for extraction, while high-level feature requires an understanding of prior complex domain-specific information to make a decision.
\end{itemize}

\section{Related work}

In recent years, LLMs have gained significant attention across the world due to their ability to complete tasks on which no explicit training was provided. The variety of applications being explored by this technology has spawned a new interest in deploying them across different domains such as medical science, business, education etc. Several studies have investigated the use of LLMs as text annotators, primarily in open domain settings \cite{ding2022gpt, he2023annollm}. However, there is limited research on evaluating their performance in applications requiring domain expertise, specifically information extraction from scientific articles \cite{dagdelen2024structured}. Studies have experimented with incorporating LLMs into data annotation pipelines, particularly for annotating texts that require domain expertise. These efforts have shown potential to reduce the time and overall cost of annotation \cite{goel2023llms}. For instance, LLMs have been used to extract information about nanorod structure procedures from scientific texts \cite{walker2023extracting} and to extract information from clinical trials \cite{ghosh2024alpapico}. 

In the context of Climate Change research, recent work such as the creation of Expert Confidence in Climate Statements (CLIMATEX) dataset \cite{lacombe2023climatex} to weigh assessment tasks in a few-shot learning setting has shown limited accuracy. However, prior to the task of assessment is the collection and streamlining of evidence from the peer reviewed literature where for example, LLMs can be useful for Named Entity Recognition (NER) \cite{mallick2024analyzing} from the vast corpus to help organize the evidence more efficiently for synthesis. Other similar trajectories in the use of LLMs in climate change research involve fact-checking of climate change claims \cite{leippold2024automated} or the utilisation of trained domain-specific climate models such as to synthesize interdisciplinary research on climate change \cite{thulke2024climategpt} or the use of LLM agents to extract information from a database to improve climate change related information analysis \cite{kraus2023enhancing}.

\begin{table*}
\centering
\fontsize{8.4}{11}\selectfont{
\begin{tabular}{p{0.1\textwidth}p{0.2\textwidth}p{0.5\textwidth}p{0.10\textwidth}}
\hline
\textbf{Feature} & \textbf{Expertise Level} & \textbf{Evidence excerpt} & \textbf{Annotation} \\
\hline
Geographic location & Low (Close to open domain standard NLP tasks) & `` ... ... purpose of this paper is to analyze how farmers are reducing vulnerability of rain-fed agriculture to drought through indigenous knowledge systems (IKS) in the Atankwidi basin, north-eastern Ghana ... ''
& Ghana  \\ 
Stakeholders & Intermediate (beyond standard tasks which requires understanding taxonomy) & ``... The results show that farmers in the Atankwidi basin are employing IKS of drought risk management for reducing vulnerability to drought in rain fed agriculture ..." &  Individuals or households \\ 
Depth & High (On-field knowledge helps discern if adaptation response is transformative) & ``... planting multiple indigenous drought resilient crop varieties and employing different rounds of seeding ... We continue to cultivate Naara and Zea because they are drought resilient with capability of surviving droughts that last a few weeks or even a month ... " & Low  \\ 
\hline
\end{tabular}
\caption{Illustrative examples of the features considered, required expertise levels for accurate extraction and corresponding annotation for illustrative evidence excerpts.}
\label{tab:examples}}
\end{table*}
\section{Method}
\subsection{Dataset}
The dataset for this study has been sourced from the Global Adaptation Mapping Initiative (GAMI) \cite{berrang2021systematic} - a global effort led by IPCC scientists to systematically collect and assess the evidence in peer-reviewed literature on climate adaptation progress. This dataset consisted of twenty five features such as Geographic Location, Adaptation Response Type, Implementation Tools etc. labeled by climate change adaptation experts from 1,682 peer-reviewed articles that met an inclusion criteria defined by the group of scientists leading the initiative. The curation of this dataset was rigorous wherein each peer-reviewed article was assigned to two human-labelers with climate change adaptation expertise and any conflict between any of the features labelled by them was resolved by a senior expert selected for their extensive experience in climate adaptation. Further details on the dataset creation can be found in the original article \cite{berrang2021systematic}.

It is important to note that the 1,682 peer-reviewed articles were divided into focus groups such as cities, food, health, etc. to help administer the labelling as well as the systematic assessment exercise. For the purposes of this study, we focus on a sample ($n=$586) of the GAMI database i.e. adaptation responses documented only in the \emph{food sector} focus group. The publications in this focus group consist information related to climate change adaptation responses aimed at ensuring food security and sustaining related livelihoods. Having worked as part of the \emph{food sector} focus group provided us easy access to the raw data. We plan to expand this analysis in the future by including data from the other focus groups. Of the twenty five features available in the dataset, we have focused on three features that reflect varying adaptation expertise to label accurately, namely: Geographic Location (Low level), Stakeholders (Intermediate level) and Depth of Adaptation Response (High level) for the sample dataset in the \emph{food sector} i.e. adaptation responses specific to the agriculture sector.
\subsection{Task Description}
To assess the utility of GPT-4o, we used three features on the sampled GAMI dataset where each feature reflects a specific level of expertise and domain knowledge in order to accurately capture information regarding the adaptation response. Table \ref{tab:examples} outlines the features categorized by their complexity based on the level of expertise needed for accurate extraction, along with illustrative examples for each feature. 
\begin{itemize}[leftmargin=*]
    \item For the low expert level we chose the extraction of geographical country where the climate change adaptation response occurred. This is similar to identifying location using open domain standard NLP tasks such as NER. Although an article may discuss multiple countries that are not relevant to the specific adaptation response in sections such as the introduction or related work, the model must accurately extract the specific location where the adaptation response occurred. This task can be viewed as classic information retrieval, and we evaluate the model using precision and recall metrics.
    \item The medium or intermediate expertise level feature is to identify the stake holder participating in the adaptation response based on a provided taxonomy: \emph{Government, Civil Society, Individuals or Households, International or multinational governance institutions,} and \emph{Private Sector}. The model was guided to identifying the stakeholders through an intermediate step in the prompt of first identifying the adaptation response discussed in the article and then to list the stakeholders involved in the said response. Classifying this requires one to be well aware of stakeholder mapping to the appropriate category which can be acquired by reading the relevant literature as well as simple training on data mapping players to their respective groups. Similar to the low expertise task, this is also an information retrieval task evaluated using precision and recall metrics.
    \item The high expertise level feature is the depth of adaptation response. The labelling for this feature requires the depth for the response to be categorised as low, medium, high or Not certain / Insufficient information / Not assessed. High depth reflects transformative changes with novel solutions. Low depth implies that the response is largely based on expansion of existing practices rather than consideration of entirely new practices. Medium depth indicates that new practices are being pursued, however they may not be transformative in nature. However, if there was a lack of clarity regarding the depth of the response the label Not certain / Insufficient information / Not assessed is chosen. In order to classify this, one would require significant expertise and understanding of the adaptation literature as well as practical on-ground experience to match the depth of the adaptation response recorded in the literature. This task is viewed as a classic multi-class classification problem, and we use precision, recall, and F1 scores for evaluation.
\end{itemize}

\subsection{Experiments}
We prompted GPT-4o\footnote{\url{https://openai.com/index/hello-gpt-4o/}} for this task by first converting the PDF files to markdown format using LlamaParse\footnote{\url{https://docs.cloud.llamaindex.ai/llamaparse/}}. To guide the model effectively, we included an intermediate verification step in the prompts. Specifically, the model was first asked to identify the climate change adaptation response and then to recognize that stakeholders involved in this response. The complete prompt used is provided in the Appendix \ref{sec:appendix}. Additionally, to understand the model's reasoning and identify differences between the model's outputs and human annotations, we asked the model to provide excerpts that it used to justify its extractions. This approach allowed us to analyze the model's rationale where it diverged from human annotations. The model was prompted under default settings.
\section{Results}
To evaluate the information extraction capabilities of GPT-4o in the context of climate change data from scientific publications, we compared the annotation agreement between human-created labels and the information extracted by GPT-4o. Table \ref{tab:results} presents a summary of the evaluation metrics, and our findings are detailed below:
\begin{table}
\centering
\fontsize{9}{11}\selectfont{
\begin{tabular}{cccc}
\hline
\textbf{Expertise} & \textbf{Precision} & \textbf{Recall} & \textbf{F1} \\
\hline
Low  & 0.88 & 0.90 & 0.89 \\
Medium & 0.40 & 0.83 & 0.54\\
High & 0.22 & 0.22 & 0.22\\
\hline
\end{tabular}
\caption{Summary of evaluation metrics showing decrease in model performance as task complexity increases.}
\label{tab:results}}
\end{table}

\noindent\textbf{Low Expertise Tasks} such as extracting the geographic regions where adaptation responses occurred, GPT-4o demonstrated high agreement with human annotators. Specifically, GPT-4o achieved a precision score of 0.88 and a recall of 0.9. In instances of disagreement, manual checks revealed that GPT-4o often provided more specific information, extracting exact countries while human annotators tended to group countries together. These results align with findings from studies that used LLMs for NER tasks, suggesting consistent performance across different domains \cite{goel2023llms}. Furthermore, this specificity indicates the potential of GPT-4o to enhance the granularity of extracted data in low-expertise tasks.

\noindent\textbf{Intermediate Expertise Tasks} of identifying stakeholders involved in adaptation responses, GPT-4o effectively captured the primary stakeholders but also extracted extraneous information and occasionally misclassified categories. The performance metrics for this level included a micro F1 score of 0.54 (macro: 0.30), precision of 0.40 (macro: 0.27), and recall of 0.83 (macro: 0.33). Manual checks of the disagreements highlighted that GPT-4o sometimes misidentified and extracted stakeholders mentioned in introductory sections or other parts of the text, which were not relevant to the specific adaptation measures being discussed in the document (high recall). Given the task requires the model to use a taxonomy to classify the stakeholders, the model's performance on this task suggests that improvements can be made by integrating prompting methods that elicit reasoning and verification capabilities. 

\noindent\textbf{High Expertise Tasks}
For high expertise feature extraction, our evaluation reveals that the model, in some cases (10.4\% of the time), provides individual assessments for each adaptation response rather than an aggregate assessment of the impact of a set of responses. This behavior complicated the evaluation process. To address this, we isolated these instances and focused our evaluation on cases where the model provided a depth evaluation for the aggregate of set of adaptation responses. We treated the evaluation as a multi-class classification problem. In this context, GPT-4o achieved an accuracy of 22.7\% and a micro-averaged F1 score of 0.22 (macro F1: 0.17). Closer examination of the instances of disagreement revealed that, in all the cases with minimal agreement, GPT-4o exhibited a more optimistic view compared to human annotators, often overestimating the impact of adaptation responses. This discrepancy is likely due to the generalized nature of the model's training and instruction tuning on a wide range of tasks. These findings highlight significant challenges in using GPT-4o for tasks that require a deep understanding of complex and nuanced information. 

\section{Limitations}
In this study, we tried to cover a diverse set of information extraction tasks in the context of climate change adaptation research to understand the feasibility of using GPT-4o, but it is in not an exhaustive list. Additionally, our findings are specific to the climate change adaptation literature and in \emph{food sector}, limiting their generalizability to other domains or sub-fields of climate change research. Further studies are necessary to assess the applicability of GPT-4o across a broader range of climate change topics. We did not explore complex prompting techniques, such as Chain of Verification \cite{dhuliawala2023chain}, which could potentially enhance the accuracy and reliability of GPT-4o outputs. Incorporating such advanced techniques in future research might address some of the challenges we encountered, such as misclassification and the extraction of irrelevant information. Our evaluation was conducted exclusively on GPT-4o, and we did not test other LLM models, which may perform differently. Future research should include a comparison of multiple LLMs to determine if our findings are consistent across different models and architectures.
\section{Conclusion}
From our study we find that there are opportunities and challenges for the deployment of pre-trained LLMs in the climate evidence synthesis and assessments. We assessed the efficiency of GPT-4o's role as an annotator and find that tasks requiring beyond low levels of expertise are challenging for GPT-4o. 
Future work should explore using methods to integrate knowledge for medium expertise level and learning from human feedback to improve the model performance on extraction on information that requires high levels of expertise. 
Further, models with such capacity when trained on task specific data, could play a complementary role in the task of adaptation tracking by governments and global agencies and eventually help in timely securing funding for necessary adaptation responses. However, it is important to emphasize that these models cannot completely replace the expert-driven process, rather a human-in-the loop system would be extremely beneficial for ensuring the integrity and effectiveness of this process.




\section*{Acknowledgements}
We would like to thank the anonymous reviewers for their helpful feedback and suggestions, which have greatly improved this paper. We also want to express our appreciation to the creators of the GAMI dataset. Their hard work in gathering and organizing the dataset was essential for our work.

\bibliography{main}
\bibliographystyle{acl_natbib}

\appendix

\section{Appendix}\label{sec:appendix}
Here we provide the prompt we have used for information extraction.
\begin{figure*}
\begin{tcolorbox}[title=Prompt,boxsep=1pt,left=3pt,right=3pt,top=3pt,bottom=3pt]
\textcolor{black}{You are a climate change research assistant with expertise in adaptation tracking through document analysis. Your task is to identify the evidence regarding the following questions below within the context of climate change adaptation:\\1. Where exactly in terms of geography is this adaptation response observed? If there are more than one location please provide all that apply. Following details, if available, must be provided in this format\\
    \hspace{2cm}Country name: <country name>, \\
    \hspace{2cm}Sub-national region: <sub national region>,\\
    \hspace{2cm}Excerpt: <Provide an excerpt from the text that justifies your selection.>\\
2. Please identify the adaptation response undertaken and for the adaptation response identified.\\
Please provide who among the following list of stakeholders (brief description of each provided in []) who are engaging with the adaptation response based on the following rubric:\\    
International or multinational governance institutions: 
[Global or regional treaty body or agency such as UN institutions/organizations, EU institutions, Organization of American States, African Union etc.],\\
 Government (national): [Countries officially recognized by the UN],\\
 Government (sub-national): [Domestic, sub-national governing unit. Terms include state, province, territory, department, canton, Lander],\\
 Government (local): [Terms include municipality, local government, community, urban, rural regions],\\
 Private sector (corporations): [Large national or international companies],\\
 Private sector (SME): [Small- and medium-enterprises],\\
 Civil society (international, multinational, national): [Voluntary civil society organizations. Includes charities, non-profits, faith-based organizations, professional organizations (e.g. labour unions, associations, federations), cultural groups, religious groups, sporting associations, advocacy groups.],\\
 Civil society (sub-national or local): [Formal community associations],\\ 
 Individuals or households: [Including informal community networks],\\
 Other: [If none of the above categories apply, please report it under "Other" and specify the entity or individual involved.] \\
 Your response for this must be in the following format:
 Stakeholders: <your answer>,\\
 Excerpt: <Please provide an excerpt from the text that justifies your selection>
 }

 Please note that the stakeholder must be involved in the adaptation response!\\

3. The depth of the climate adaptation response relates to the degree to which a change reflects something new, novel, and different from existing norms and practices. \\
A change that has limited depth would follow business-as-usual practices, with no real difference in the underlying values, assumptions and norms. \\
This would include responses that are largely based on expansion of existing practices rather than consideration of entirely new practices. In-depth change, in contrast, might involve radically changing practices by altering frames, values, logics, and assumptions underlying the system.\\
This might involve deep structural reform, complete change in mindset by governments or populations, radical shifts in public perceptions or values, and changing institutional or behavioral norms. \\
Based on your assessment classify the depth of the adaptation response identified as any of the following: Low; Medium; High; Not certain / Insufficient information / Not assessed. \\

Your response for this must be in the following format:

\hspace{2cm} Depth: <your assessment>,

\hspace{2cm} Explanation: <your reasoning for this assessment>\\

Here is the document in markdown format: \{document\}
\end{tcolorbox}
\end{figure*}

\end{document}